\documentclass[11pt, a4paper, copyright, nonumbering]{include/style}

\usepackage[authoryear, sort&compress, round]{natbib}

\usepackage{graphicx}
\usepackage{subcaption}

\usepackage{multirow}
\usepackage{tabularx}
\usepackage{makecell}
\usepackage[table]{xcolor}
\definecolor{lightgray}{gray}{0.9}

\newcolumntype{Y}{>{\hspace{3pt}}X<{\hspace{3pt}}}

\usepackage{mathtools}
\usepackage{dsfont}

\usepackage{algorithm}
\usepackage{algorithmic}

\usepackage{siunitx}
\usepackage[capitalize,noabbrev]{cleveref}

\usepackage{fontawesome5} 
\usepackage[most]{tcolorbox}
\usepackage{CJKutf8}
\usepackage{placeins}

\newtcolorbox{promptbox}[2][]{%
  colback=#2!5,           % box background
  colframe=#2!60,         % border color
  colbacktitle=#2!70,     % title background
  coltitle=black,         % title text color
  boxrule=0.4pt,
  arc=3mm,
  enhanced,
  left=2mm,right=2mm,top=1ex,bottom=1ex,
  fonttitle=\bfseries,
  title=#1,
}

\title{\centering Constraint-Rectified Training for Efficient Chain-of-Thought}
\date{}
\author[*]{
Qinhang Wu$^*$$^1$, Sen Lin$^2$, Ming Zhang$^3$, Yingbin Liang$^1$, Ness B.~Shroff$^1$ \\
\small $^1$The Ohio State University~~~ $^2$University of Houston~~~ $^3$Google \\
}
\begin{abstract}
Chain-of-Thought (CoT) has significantly enhanced the reasoning capabilities of Large Language Models (LLMs), especially when combined with reinforcement learning (RL) based post-training methods. While longer reasoning traces can improve answer quality and unlock abilities such as self-correction, they also incur high inference costs and often introduce redundant steps, known as overthinking. Recent research seeks to develop efficient reasoning strategies that balance reasoning length and accuracy, either through length-aware reward design or prompt-based calibration. However, these heuristic-based approaches may suffer from severe accuracy drop and be very sensitive to hyperparameters. To address these problems, we introduce \textbf{CRT} (\textbf{C}onstraint-\textbf{R}ectified \textbf{T}raining), a principled post-training framework based on reference-guarded constrained optimization, yielding a more stable and interpretable formulation for efficient reasoning without fragile threshold tuning. CRT alternates between minimizing reasoning length and rectifying accuracy only when performance falls below the reference, enabling stable and effective pruning of redundant reasoning. We further extend CRT with a two-stage training scheme that first discovers the shortest reliable reasoning patterns and then refines accuracy under a learnt length budget, preventing the re-emergence of verbose CoT. Our comprehensive evaluation shows that this framework consistently reduces token usage while maintaining answer quality at a robust and reliable level. Further analysis reveals that CRT improves reasoning efficiency not only by shortening responses but also by reducing internal language redundancy, leading to a new evaluation metric. Moreover, CRT-based training naturally yields a sequence of intermediate checkpoints that span a spectrum of explanation lengths while preserving correctness, enabling fine-grained control over reasoning verbosity without retraining.

\end{abstract}
\begin{document}

\begin{CJK*}{UTF8}{gbsn}

\maketitle

\begingroup
\renewcommand\thefootnote{}
\footnotetext{*Correspondence to Qinhang Wu (wu.5677@osu.edu).}
\endgroup

\section{Introduction}

Frontier reasoning models such as OpenAI o1 \citep{jaech2024openai} and Gemini 2.5 \citep{comanici2025gemini} have demonstrated remarkable proficiency in addressing complex tasks in mathematics, programming, and logical reasoning across diverse domains. Meanwhile, open-source families such as Qwen3 and Phi-4 series provide a spectrum of parameter scales, substantially advancing research in language model reasoning. At the heart of these successes lies Chain-of-Thought (CoT), which allows models to reason step by step and achieve higher performance on challenging problems.

While CoT strengthens reasoning capability, it frequently produces redundant outputs, a phenomenon termed \textit{overthinking} \citep{sui2025stop}, which incurs computational overhead and can even degrade answer quality. Efforts to address this \textit{efficient reasoning} problem primarily span supervised fine-tuning (SFT), inference-time add-ons, and post-training reinforcement learning. Among them, post-training RL is especially powerful since it directly optimizes accuracy and efficiency through reward shaping, rather than relying on prompt curation or decoding strategies. However, many existing approaches rely on heuristic reward design or fixed accuracy thresholds that are difficult to tune and often lead to unstable tradeoffs between correctness and brevity.

This naturally leads to the following question: \emph{Can we develop a principled approach to enhance reasoning efficiency without sacrificing accuracy?} In this work, we cast efficient reasoning as a constrained optimization problem, in which the reasoning length is minimized subject to constraints that ensure reasoning accuracy is maintained. More specifically, our contributions can be summarized as follows:
\begin{itemize}
    \item We formulate efficient reasoning as a principled \emph{reference-guarded constrained optimization} problem,
    where a relative safeguard against a frozen reference policy is introduced to replace fragile absolute-value based accuracy constraints.
    \item We introduce \textbf{Constraint-Rectified Training (CRT)}, a simple and principled post-training algorithm that alternates between shortening reasoning traces and restoring accuracy only when performance falls below the reference, enabling stable and effective pruning of redundant reasoning.
    \item We further extend CRT with a two-stage training scheme that first discovers the shortest reliable reasoning patterns and then refines accuracy under a fixed length budget, preventing the re-emergence of verbose CoT. 
    \item We conduct comprehensive evaluations across multiple reasoning benchmarks, demonstrating that CRT consistently reduces reasoning redundancy while maintaining robust answer accuracy. Beyond average length, we further propose a quantitative metric to measure internal language redundancy, revealing efficiency gains that are not captured by token count alone.
\end{itemize}
\begin{figure}[ht]
  \centering
  \hspace{3mm}
  \includegraphics[width=0.95\linewidth]{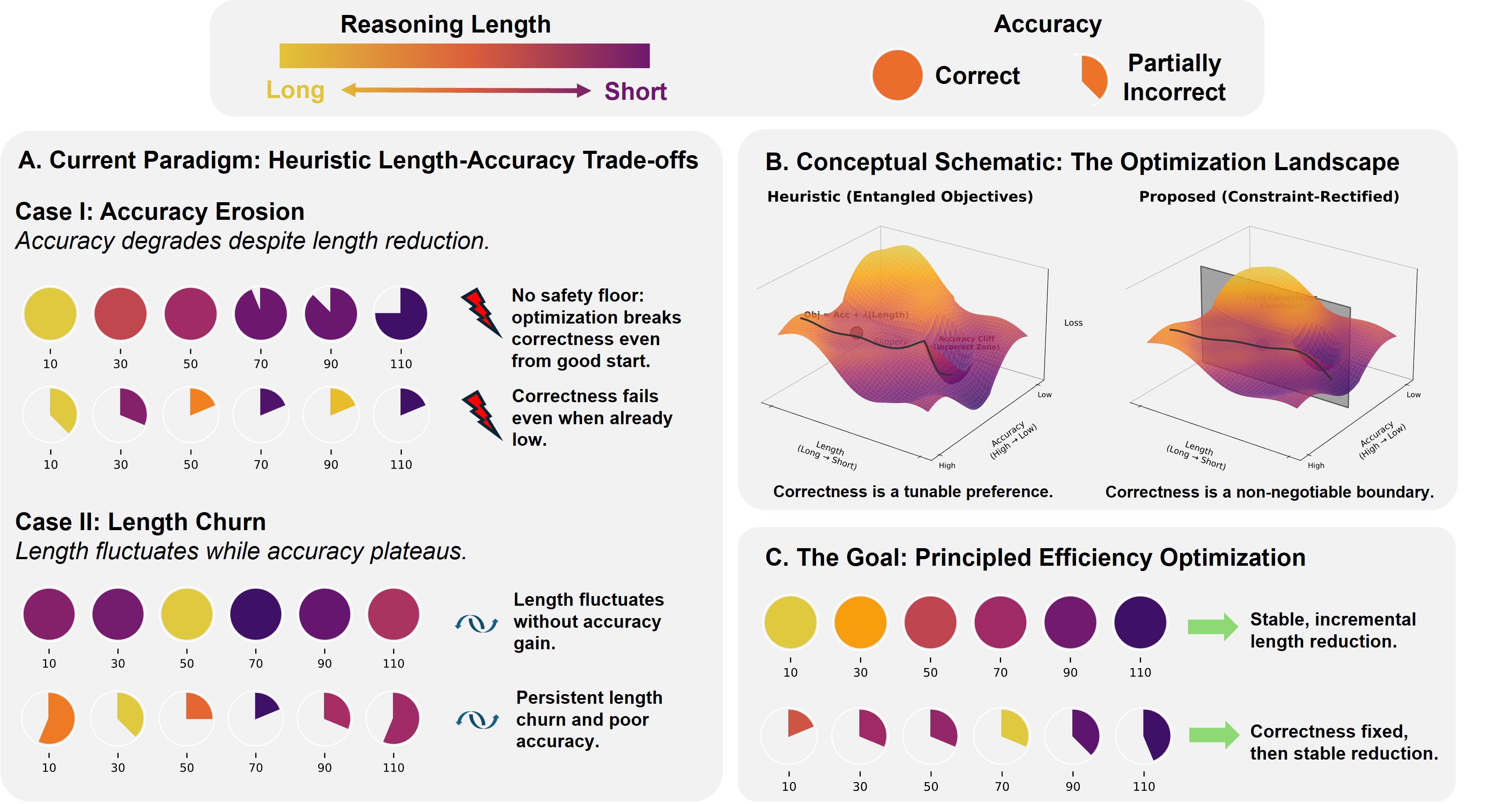}
  \caption{Failure modes of heuristic length-accuracy tradeoffs. Color intensity encodes the response length (brighter = longer; darker = shorter). Circle fill represents accuracy: a fully filled circle marks a correct prediction, whereas a partially filled circle reflects partial correctness, with larger filled regions corresponding to higher correctness. Numbers below the circles denote training steps.}
  \label{fig:motivation}
\end{figure}
\section{Related Work}

\textbf{RL-based reasoning}. CoT prompting \citep{wei2022chain} shows that inducing intermediate steps can substantially improve  model's performance on a wide range of reasoning tasks. Subsequent extensions enhance CoT by reasoning traces aggregation \citep{wang2022self} and tree-based CoT generalization \citep{yao2023tree}. RL has recently re-emerged as a practical recipe for training Large Reasoning Model (LRM) \citep{cobbe2021training,lightman2023let,luo2023wizardmath,wang2024math,zeng2025simplerl,zhu2025retrieval}, popularized by the renewed adoption of RL-based post-training in DeepSeek-R1 \citep{guo2025deepseek}. A representative approach is GRPO \citep{shao2024deepseekmath}, which replaces an explicit reward model with group-relative comparisons among multiple rollouts per prompt; this line has  been extended by many variants \citep{wu2025takes,zhang2025scaf,mansouri2025noise,yu2025dapo}. In parallel, a growing body of work characterizes reasoning traces as structured objects to better understand how post-training shapes reasoning behavior \citep{chen2025your,hu2025llms,singla2025thinking}.

Despite the apparent logical form of CoT, reasoning traces are associated with a diverse set of failures beyond final-answer correctness. Recent studies highlight \emph{illusionary} or invalid reasoning, where the trace may appear plausible without faithfully supporting the answer \citep{weatherhead2025illusions,he2025makes,jose2025reasoning,liu2025verifying}; \emph{overthinking}, where redundant steps inflate cost and can even degrade accuracy \citep{chen2024not}; \emph{underthinking}, where LRMs still fail to allocate sufficient effective computation \citep{wangthoughts}; \emph{collapse} into illegible or degenerate chains \citep{jose2025reasoning}; \emph{forgetting} induced by reasoning-centric post-training \citep{phan2025beyond}; and \emph{data contamination} that can confound conclusions about genuine reasoning improvements \citep{wang2025fragility,dang2025memories}. These findings collectively call for training frameworks that explicitly regulate how computation is allocated during reasoning.

\textbf{Overthinking}. The efficiency cost of reasoning has attracted particular attention: unnecessary CoT steps increase latency and token usage, and may introduce additional opportunities for error. Existing approaches mitigate overthinking from several angles, including supervised shortening or compression of reasoning traces \citep{kang2025c3ot,xia2025tokenskip,liu2024can}, RL-based post-training that directly optimizes for efficient reasoning \citep{yi2025shorterbetter,luo2025o1,hou2025thinkprune,arora2025training,shen2025dast,liu2025bingo,li2025aalc}, and inference-time controls such as early exiting based on intermediate signals \citep{dai2025s,hao2024training}. However, many methods still hinge on heuristic thresholds or sensitive trade-offs between brevity and correctness, leaving open the question of how to enforce efficiency while preserving answer reliability in a principled manner.

\section{Motivation}

Despite the empirical success of CoT reasoning, inefficiency remains a critical issue for practical deployment, particularly consequential in budget-constrained environments. Existing approaches predominantly quantify inefficiency through response length, implicitly assuming that shorter reasoning traces correspond to more efficient reasoning. However, we argue that reasoning inefficiency is not merely a matter of verbosity, but reflects deeper optimization pathologies that are poorly addressed by pure length-based objectives.

From an optimization perspective, CoT post-training involves competing objectives: improving task accuracy while reducing reasoning cost. Most existing approaches address this tension by collapsing efficiency into a scalar proxy such as response length and heuristically combining it with an accuracy-based reward. While simple and intuitive, this design treats correctness as a tunable preference rather than an explicit constraint, obscuring its structural role in the optimization problem. 

The consequences of this entanglement are illustrated in \cref{fig:motivation}. Panel A summarizes two failure modes observed during training with heuristic length-accuracy tradeoffs. In \textbf{accuracy erosion}, models achieve shorter responses at the cost of degraded correctness, even when starting from a strong baseline. In \textbf{length churn}, response length fluctuates substantially across training iterations while accuracy remains largely unchanged, indicating inefficient optimization that fails to suppress redundant reasoning. These behaviors reflect unstable optimization trajectories rather than isolated hyperparameter choices.

These failure modes are not merely incidental. \cref{table:motivation} quantifies accuracy stability conditioned on successful length reduction (Len$\downarrow$). Across both datasets, selected heuristic baselines \textit{ThinkPrune} and \textit{ACPO} frequently exhibit accuracy degradation when length decreases: up to one-third of length-reduced instances suffer correctness loss. 
In contrast, CRT exhibits substantially higher accuracy preservation under Len$\downarrow$, with a larger fraction of cases maintaining or improving correctness. This gap highlights a fundamental limitation of length-based approaches: since correctness is not explicitly enforced, they provide no mechanism to bound performance degradation. Small changes in length pressure can therefore induce unwanted accuracy shifts.

\begin{table}[ht]
\footnotesize
\centering
\caption{Accuracy stability under successful length reduction. Each entry reports the percentage of problems whose accuracy decreases (AD\%), perserves (AP\%), or improves (AI\%) after training, conditioned on instances where response length is reduced.}
\label{table:motivation}
\begin{tabularx}{\columnwidth}{lXXXX}
\specialrule{.2em}{.1em}{.25em}
\textbf{Method} & AD\% ($\downarrow$) & AP\% ($\uparrow$) & AI\% ($\uparrow$) & Acc ($\uparrow$)  \\
\midrule
\multicolumn{4}{l}{\textbf{SAT Math}} \\
    \textit{ThinkPrune-1k} & 37.5\% & 45.8\% & 16.7\% & 81.45 \\
    \textit{ACPO} & 27.3\% & 40.9\% & 31.8\% & 91.02 \\
    \rowcolor{blue!8} \textit{CRT (ours)} & 12.5\% & 56.2\% & 31.2\% & 93.16 \\
\midrule
\multicolumn{4}{l}{\textbf{GSM8K}} \\
    \textit{ThinkPrune-1k} & 34.0\% & 46.1\% & 19.9\% & 81.54 \\
    \textit{ACPO} & 28.7\% & 51.1\% & 20.2\% & 82.78 \\
    \rowcolor{blue!8} \textit{CRT (ours)} & 20.0\% & 55.6\% & 24.4\% & 84.64 \\
\specialrule{.2em}{.1em}{.1em}
\end{tabularx}
\end{table}

\paragraph{Toward Principled Efficiency Optimization.}
The insights above indicate that reasoning inefficiency is not merely a matter of verbosity, but a consequence of how efficiency and correctness are coupled during optimization. This motivates a shift from heuristic reward mixing toward a constraint-aware formulation. We advocate for explicitly rectifying efficiency optimization with correctness constraints. Such formulation cleanly separates \textit{what must not be sacrificed} (correctness) from \textit{what should be improved} (reasoning efficiency), yielding three key advantages:
\begin{enumerate}
    \item Principled guarantees: correctness is preserved reliably rather than recovered post hoc;
    \item Stable optimization dynamics: redundant reasoning patterns are suppressed even when they do not immediately affect accuracy;
    \item Fine-grained accuracy control: efficiency can be optimized incrementally without brittle threshold tuning.
\end{enumerate}

\section{Method}

In this section, we introduce \emph{Constraint-Rectified Training} (CRT), a post-training framework that enables systematic pruning of redundant reasoning while maintaining robust correctness.

\subsection{Preliminaries}

Let $x$ denote an input prompt drawn from a distribution $\rho$, and let $y$ represent the model-generated output containing both CoT reasoning and the final answer. We denote $y^*(x)$ as the ground-truth answer corresponding to prompt $x$, and model responses are sampled from a policy $\pi_{\theta}(y \mid x)$ parameterized by~$\theta$. To quantify efficiency, we introduce a normalized length measure $\ell_{\mathrm{norm}}(y)$, obtained by normalizing response length with respect to the per-prompt mean and variance, yielding a scale-invariant length measure. Solution correctness is evaluated using a verifier-based indicator $\mathds{1}\{y = y^*(x)\}$.

\subsection{Formulation}

We formulate \emph{efficient reasoning} as a constrained optimization problem: the goal is to reduce the expected length of generated solutions while ensuring that accuracy does not fall below a specified threshold $\alpha$. Formally, we optimize
\begin{equation}
\begin{aligned}
\min_{\pi} \quad & \mathbb{E}_{x \sim \rho,\, y \sim \pi_\theta(y \mid x)} \left[ \ell_{\mathrm{norm}}(y) \right] \\
\text{s.t.} \quad & -\mathbb{E}_{x \sim \rho,\, y \sim \pi_\theta(y \mid x)} \left[ \mathds{1}\{y = y^*(x)\} \right] \leq \alpha
\end{aligned}
\label{eq:formulation1}
\end{equation}
which captures the central tension in efficient reasoning: shortening CoT without sacrificing correctness beyond an acceptable margin. 

However, it is often impractical to specify an appropriate accuracy threshold $\alpha$ before post-training for  ~\cref{eq:formulation1}. If the threshold is set too high, the feasible region becomes overly restrictive and the model cannot meaningfully shorten its reasoning. Conversely, if the threshold is set too small, the optimization may aggressively prune CoT steps, leading to severe degradation in reasoning quality. \footnote{In the extreme case, the model may omit intermediate reasoning entirely and directly output a final answer.} In short, the scalar constraint parameter $\alpha$ is difficult to tune and does not provide a stable accuracy safeguard during training. 
To address this problem, 
we introduce a \emph{reference-policy guard} to replace the fixed accuracy threshold. Let $\pi_{\mathrm{ref}}$ denote a frozen reference model (e.g., the initial model before any length-aware post-training). Since $\pi_{\mathrm{ref}}$ represents the original reasoning capability of the model, an intuitive requirement is that our updated policy $\pi$ should not perform significantly worse than $\pi_{\mathrm{ref}}$ in terms of correctness. This leads to a relative accuracy constraint:
\begin{equation}
\begin{aligned}
&\mathbb{E}_{x \sim \rho,\, y_{\mathrm{ref}} \sim \pi_{\mathrm{ref}}(y \mid x)}
\!\left[\mathds{1}\{y = y^*(x)\}\right]
\;\ge\; \mathbb{E}_{x \sim \rho,\, y_{\mathrm{ref}} \sim \pi_{\mathrm{ref}}(y \mid x)}  
\!\left[\mathds{1}\{y_{\mathrm{ref}} = y^*(x)\}\right]
- \varepsilon
\end{aligned}
\end{equation}
where $\varepsilon \ge 0$ is a small tolerance indicating how much accuracy degradation is allowed. This relative constraint avoids the difficulty of selecting an absolute accuracy threshold and ensures that length pruning does not undermine the model's original problem-solving performance.
Under this reference-based accuracy safeguard, the objective can be reformulated as follows:
\begin{equation}
\begin{aligned}
\min_{\pi} \quad &
\mathbb{E}_{x \sim \rho,\, y \sim \pi_\theta(y \mid x)}
\!\left[ \ell_{\mathrm{norm}}(y) \right] \\[2pt]
\text{s.t.} \quad &
\mathbb{E}_{x \sim \rho,\, y \sim \pi_\theta(y \mid x)}
\!\left[ \mathds{1}\{y = y^*(x)\} \right]  \ge
\mathbb{E}_{x \sim \rho,\, y_{\mathrm{ref}} \sim \pi_{\mathrm{ref}}(y \mid x)}
\!\left[ \mathds{1}\{y_{\mathrm{ref}} = y^*(x)\} \right]
- \varepsilon .
\end{aligned}
\label{eq:formulation2}
\end{equation}

\subsection{Constraint-Rectified Training}

We now turn to developing a practical post-training method that can enforce this constraint during optimization. A natural first attempt to enforce the above constraint is to adopt a primal-dual optimization framework, in which the accuracy constraint is incorporated via a Lagrange multiplier and jointly optimized with the primary objective. While conceptually straightforward, such approaches often introduce additional sensitivity to hyperparameter tuning, as the dual variable must be carefully calibrated to balance accuracy preservation and length reduction. In practice, we observe that improper tuning can lead to oscillatory behavior or overly conservative updates, limiting the effectiveness of reasoning pruning.

To avoid the use of dual variables, in this paper we leverage the Constraint-Rectified Policy Optimization (CRPO) framework~\citep{xu2021crpo}, which performs immediate, lightweight switches between minimizing length and maintaining accuracy. This aligns directly with our goal: to prune unnecessary reasoning while ensuring the model does not degrade relative to the reference policy.
Under our updated formulation, we modify CRPO so that accuracy is enforced relative to the reference model rather than via an absolute threshold. The algorithm therefore alternates between (i) shortening reasoning length when accuracy is sufficiently high, and (ii) restoring accuracy whenever the model underperforms the reference. We propose the resulting algorithm as CRT detailed in \cref{algo:main-crt}. The procedure is simple, stable in practice, and naturally implements our reference-guarded constraint.

\begin{algorithm}[ht]
\caption{Constraint-Rectified Training}
\small
\label{algo:main-crt}
\begin{algorithmic}[1]
\STATE \textbf{Input:} Initial LM parameters $\theta_0$, reference policy $\pi_{\mathrm{ref}}$, tolerance $\varepsilon$, CRPO slack $\eta>0$, learning rate $\eta_\theta$, total steps $T$.
\FOR{step $t = 0,1,\dots,T-1$}
    \STATE Sample minibatch $\{x_i\}_{i=1}^B \sim \rho$, generate $y_i \sim \pi_\theta(\cdot \mid x_i)$
    \STATE Generate reference responses $y_i^{\mathrm{ref}} \sim \pi_{\mathrm{ref}}(\cdot \mid x_i)$
    \STATE Evaluate minibatch accuracy of current and reference policies:
    \[
    \hat{A}_t = \frac{1}{B}\sum_{i=1}^B \mathds{1}\{y_i = y^*(x_i)\},
    \quad
    \hat{A}^{\mathrm{ref}} = \frac{1}{B}\sum_{i=1}^B \mathds{1}\{y_i^{\mathrm{ref}} = y^*(x_i)\}
    \]
    \IF{$\hat{A}_t < \hat{A}^{\mathrm{ref}} - \varepsilon - \eta$}
        \STATE /* accuracy below reference → rectify toward accuracy */
        \[
        \theta_{t+1} \leftarrow \theta_t + \eta_\theta \nabla_\theta 
        \left( \frac{1}{B}\sum_{i=1}^B \mathds{1}\{y_i = y^*(x_i)\} \right)
        \]
    \ELSE
        \STATE /* accuracy acceptable → optimize for shorter reasoning */
        \[
        \theta_{t+1} \leftarrow \theta_t - \eta_\theta \nabla_\theta 
        \left( \frac{1}{B}\sum_{i=1}^B \ell_{\mathrm{norm}}(y_i) \right)
        \]
    \ENDIF
\ENDFOR
\end{algorithmic}
\end{algorithm}

\subsection{Two-stage Extension of CRT}
The reference-guarded formulation in \cref{eq:formulation2} enables effective pruning of redundant reasoning by minimizing response length while ensuring that accuracy does not fall below a fixed reference policy. This design is well aligned with our primary objective of obtaining concise yet reliable reasoning trajectories. However, after Stage~I converges, the policy usually reaches a saturated regime where the reasoning length can no longer be substantially reduced. Beyond this point, further updates yield limited improvements in length, while gains in accuracy become comparatively more meaningful and valuable.

Thus motivated, we introduce a novel two-stage extension of CRT that decouples efficiency discovery from accuracy refinement. Stage~I aggressively compresses reasoning traces under reference-guarded correctness constraints. Stage~II then reverses the roles of objective and constraint, but crucially, it avoids introducing an absolute length target which would again require brittle threshold selection. Instead, we use the Stage~I trained policy itself as a length reference, and constrain Stage~II to match the normalized length profile achieved by that Stage~I reference model. This modification yields a principled accuracy-refinement phase without sacrificing the efficiency discovered in Stage~I.

\paragraph{Stage~I: Length Minimization under Reference-Guarded Accuracy Constraints.} In the first stage, we solve the reference-guarded constrained optimization problem in \cref{eq:formulation2}, producing a compressed policy whose correctness remains competitive with that of the original reference policy. Once improvement in response length saturates, we obtain the Stage~I reference policy, denoted by $\pi_{\mathrm{ref}}^{(I)}$.

\paragraph{Stage~II: Accuracy Maximization under Reference-Based Length Constraints.}
In the second stage, we shift the optimization focus to maximizing solution correctness while constraining the expected normalized length to remain close to that of the Stage~I policy $\pi_{\mathrm{ref}}^{(I)}$:
\begin{equation}
\begin{aligned}
\max_{\pi} \quad &
\mathbb{E}_{x \sim \rho,\, y \sim \pi_\theta(y \mid x)}
\!\left[ \mathds{1}\{y = y^*(x)\} \right] \\[2pt]
\text{s.t.} \quad &
\mathbb{E}_{x \sim \rho,\, y \sim \pi_\theta(y \mid x)}
\!\left[ \ell_{\mathrm{norm}}(y) \right] \le
\mathbb{E}_{x \sim \rho,\, y_{\mathrm{ref}}^{(I)} \sim \pi_{\mathrm{ref}}^{(I)}(y \mid x)}
\!\left[ \ell_{\mathrm{norm}}\!\left(y_{\mathrm{ref}}^{(I)}\right) \right]
+ \delta ,
\end{aligned}
\label{eq:stage2_ref}
\end{equation}
where $\delta \ge 0$ is a small tolerance controlling how much the policy is allowed to deviate (in expectation) from the Stage~I length regime. To ensure a consistent length scale across stages, we freeze the statistics used in length normalization (e.g., mean and variance) based on rollouts from $\pi_{\mathrm{ref}}^{(I)}$.

This second stage prevents the model from reintroducing unnecessarily long CoT traces to recover accuracy and preserves the learned efficiency profile discovered in Stage~I, while encouraging the policy to find higher-quality reasoning strategies within that constraint.

\section{Experiments}
\subsection{Experimental Setup}

We implement our method within the Verl \citep{sheng2025hybridflow} RL framework. All experiments are conducted using the hardware configuration detailed in \cref{sec:appendix-hardware}. We evaluate our approach on two backbone language models: DeepSeek-R1-Distilled-Qwen-1.5B \citep{guo2025deepseek} (hereafter \textit{DeepSeek-1.5B}), and \textit{Qwen3-4B} \citep{qwen3}. Across all settings, we use a fixed batch size of $128$ and generate $16$ rollouts per prompt during training and testing.

\textbf{Dataset.} Following SimpleRL-Zoo \citep{zeng2025simplerl}, we perform RL on a mixture of the training splits from GSM8K \citep{cobbe2021training} and MATH \citep{hendrycks2021measuring}. GSM8K consists of $7.4k$ training and $1.3k$ test examples, while MATH provides $7.5k$ training and $5k$ test examples. These datasets are well-suited for studying reasoning efficiency for several reasons. First, they exhibit complementary difficulty regimes: GSM8K emphasizes elementary, multi-step arithmetic reasoning, whereas MATH contains more challenging problems requiring deeper symbolic manipulation. Second, both datasets provide standardized, in-domain evaluation splits with relatively low inference cost, enabling controlled and systematic analysis of reasoning behaviors under RL training.
For evaluation, we consider both in-domain and out-domain benchmarks. In-domain performance is assessed on GSM8K \citep{cobbe2021training} and  MATH500 \citep{lightman2023let}, which follow the same problem distributions as the RL training data. To evaluate generalization beyond the training domain, we additionally report results on a suite of out-domain math reasoning benchmarks, including SAT Math \citep{zhong2023agieval}, AMC23 \citep{hf_ds_amc23}, AIME24 \citep{aime2024}, and OLYMPIAD Bench \citep{he2024olympiadbench}.

\textbf{Baselines.} In addition to the two base models, we compare our proposed method (denoted as CRT) against reproduced results from recent approaches for efficient reasoning, including ThinkPrune \citep{hou2025thinkprune}, L1 \citep{aggarwal2025l1}, O1-Pruner \citep{luo2025o1}, ShorterBetter \citep{yi2025shorterbetter}, and ACPO \citep{acpo}. For ThinkPrune, we evaluate two length budgets, $1k$ and $500$. For O1-Pruner, we test two penalty coefficients: $\lambda \in \{1.0,4.0\}$.

\textbf{Metrics.} We report \textbf{Acc}, defined as the pass@1 accuracy (in $\%$) averaged over rollouts per question, and \textbf{Len}, defined as the average number of generated tokens per response. In addition, to better capture the trade-off between reasoning quality and efficiency, we report two composite metrics: \textbf{AES$_1$}, and \textbf{AES$_2$}. Higher AES  \citep{luo2025o1} values indicate more favorable efficiency-accuracy trade-off, with AES$_2$ imposing a stricter penalty on accuracy degradation than AES$_1$. The formal definition of AES is provided in \cref{eq:ax-aes}.

\subsection{Main Results}

\begin{figure}[t]
\centering
\begin{minipage}[t]{0.69\linewidth}
    \centering
    \includegraphics[width=\linewidth]{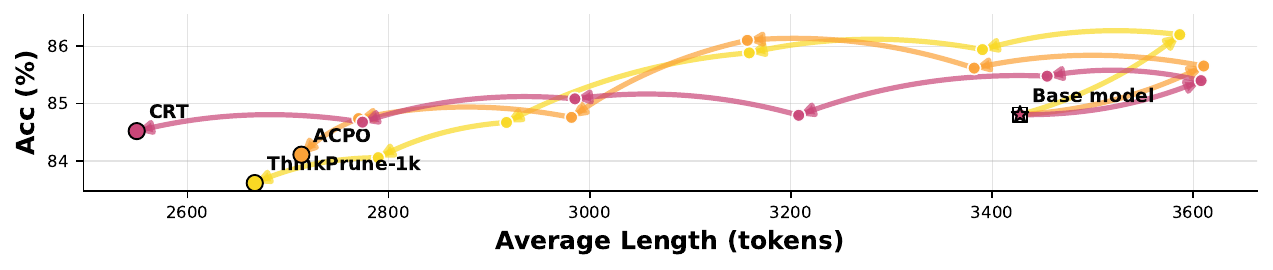}\par\vspace{0.6em}
    \includegraphics[width=\linewidth]{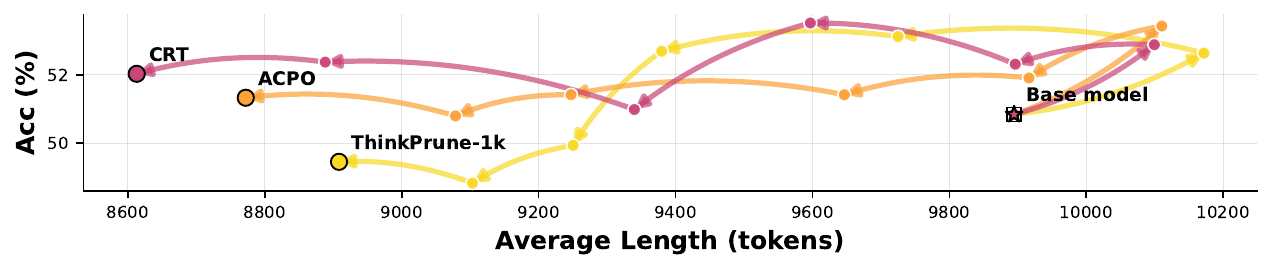}
\end{minipage}
\begin{minipage}[c]{0.28\linewidth}
    \centering
    \vspace{0.5\baselineskip}
    \includegraphics[width=1.1\linewidth]{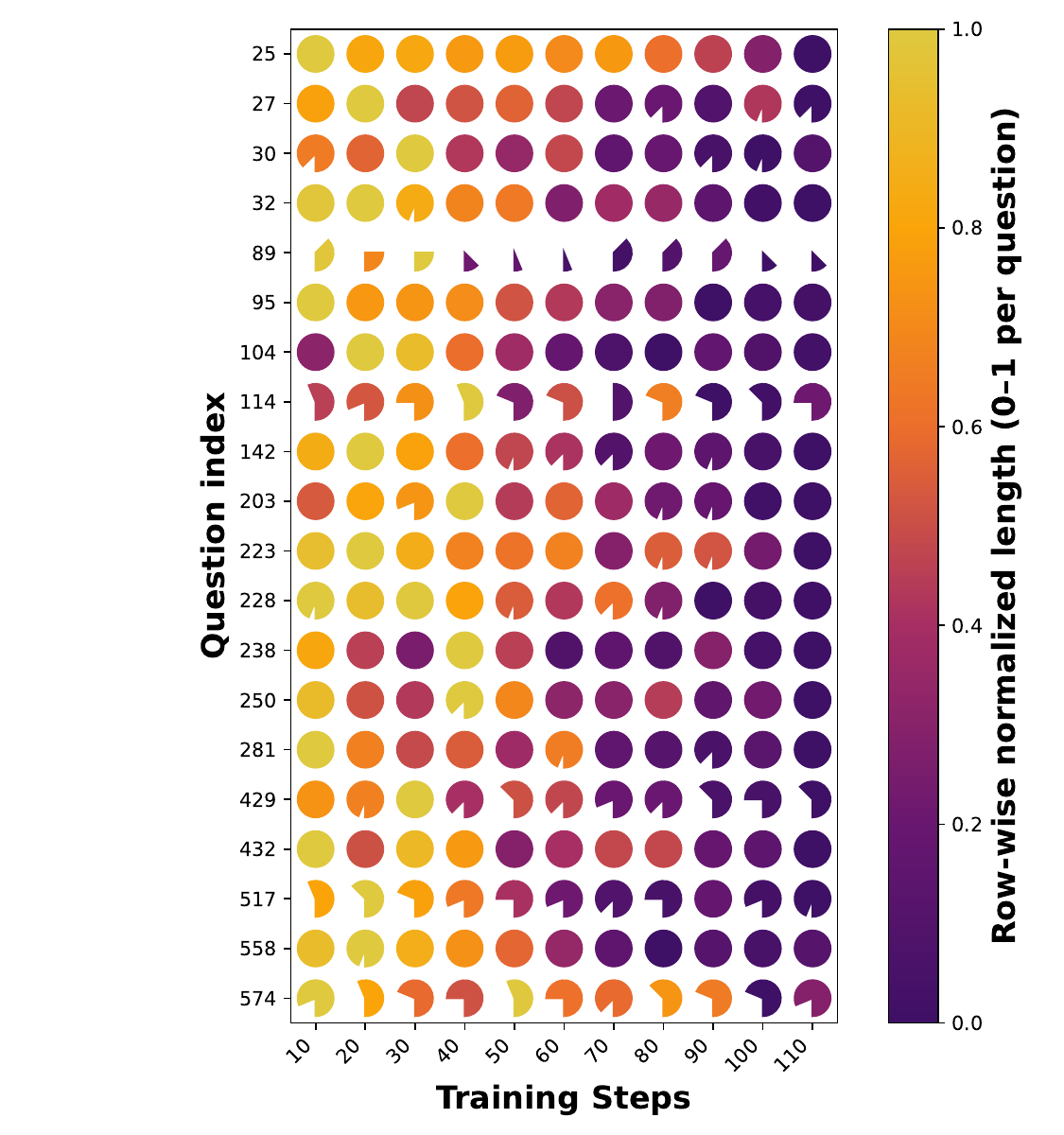}
\end{minipage}
    \caption{Left: Inference trajectories over training checkpoints. We periodically save intermediate model checkpoints during training and evaluate them via inference on the testing dataset. The base model is  DeepSeek-1.5B. The upper figure reports in-domain performance, while the lower figure reports out-domain performance. Right: Case study of CRT training behavior evaluated on GSM8K. Each row corresponds to a question, and each column to an intermediate model checkpoint generated during CRT training.}
    \label{fig:main-training-ckp}
\end{figure}

\begin{table}[h]
\footnotesize
\centering
\caption{Empirical evidence on in-domain testing dataset: GSM8K and MATH500. }
\label{table:main-1}
\begin{tabularx}{\textwidth}{lXXXXXXXXXXX}
\specialrule{.2em}{.1em}{.25em}
\textbf{Method} & \multicolumn{2}{c}{\textbf{GSM8K}} & \multicolumn{2}{c}{\textbf{MATH500}} & \multicolumn{2}{c}{\textbf{Avg}} & \multicolumn{2}{c}{\textbf{Metrics}} \\
\cmidrule(lr){2-3} \cmidrule(lr){4-5} \cmidrule(lr){6-7}  \cmidrule(lr){8-9} 
 & Acc & Len & Acc & Len & $\overline{A}$ & $\overline{L}$ & AES$_1$ & AES$_2$ \\
\midrule
    \textbf{DeepSeek-1.5B} & 84.43 & 1869 & 85.19 & 4987 & 84.81 & 3428.0 & 0.0000 & 0.0000 \\    \textit{+ ThinkPrune-500} & 78.91 & 995 & 85.25 & 4455 & 82.08 & 2725.0 & 0.0441 & -0.1168 \\
    \textit{+ ThinkPrune-1k} & 81.54 & 1128 & 85.69 & 4206 & 83.62 & 2667.0 & 0.1515 & 0.0811 \\
    \textit{+ L1} & 86.34 & 2272 & 85.69 & 4891 & 86.02 & 3581.5 & -0.0022 & -0.0022 \\
    \textit{+ O1-Pruner-1.0} & 80.36 & 1272 & 85.75 & 4174 & 83.06 & 2723.0 & 0.1022 & -0.0013 \\
    \textit{+ O1-Pruner-4.0} & 84.39 & 1561 & 85.56 & 4452 & 84.97 & 3006.5 & 0.1288 & 0.1288 \\
    \textit{+ ShorterBetter}& 71.24 & 294 & 82.47 & 3205 & 76.85 & 1749.6 & 0.0207 & -0.4482 \\
    \textit{+ ACPO} & 82.78 & 1252 & 85.44 & 4175 & 84.11 & 2713.3 & 0.1673 & 0.1261 \\
    \rowcolor{blue!8} \textit{+ CRT (ours)} & 84.64 & 1172 & 86.06 & 3826 & 85.35 & 2499.2 & 0.2901 & 0.2901 \\
\midrule
\specialrule{.2em}{.1em}{.1em}
\end{tabularx}
\end{table}

\begin{table}[h]
\footnotesize
\centering
\caption{Empirical evidence on out-domain testing dataset: SAT Math, AMC23, AIME24, and OLYMPIAD Bench. }
\label{table:main-2}
\begin{tabularx}{\textwidth}{lXXXXXXXXXXXXXXX}
\specialrule{.2em}{.1em}{.25em}
\textbf{Method} & \multicolumn{2}{c}{\textbf{SAT}} & \multicolumn{2}{c}{\textbf{AMC23}} & \multicolumn{2}{c}{\textbf{AIME24}} & \multicolumn{2}{c}{\textbf{Olymp.}} & \multicolumn{2}{c}{\textbf{Avg}} & \multicolumn{2}{c}{\textbf{Metrics}} \\
\cmidrule(lr){2-3} \cmidrule(lr){4-5} \cmidrule(lr){6-7} \cmidrule(lr){8-9} \cmidrule(lr){10-11}  \cmidrule(lr){12-13} 
 & Acc & Len & Acc & Len & Acc & Len & Acc & Len & $\overline{A}$ & $\overline{L}$ & AES$_1$ & AES$_2$  \\
\midrule
    \textbf{DeepSeek-1.5B} & 89.84 & 1373 & 69.84 & 8634 & 30.42 & 14044 & 13.25 & 15527 & 50.84 & 9894.5 & 0.00 & 0.00 \\
    \textit{+ ThinkPrune-500} & 81.45 & 609 & 71.88 & 7715 & 27.92 & 13378 & 13.63 & 14791 & 48.72 & 9123.2 & -0.13 & -0.33 \\
    \textit{+ ThinkPrune-1k} & 81.45 & 759 & 72.97 & 7548 & 28.96 & 13001 & 14.44 & 14326 & 49.46 & 8908.5 & -0.03 & -0.17  \\
    \textit{+ L1}  & 97.66 & 2377 & 72.34 & 8134 & 29.38 & 13565 & 12.56 & 15533 & 52.98 & 9902.2 & 0.12 & 0.12  \\
    \textit{+ O1-Pruner-1.0} & 86.91 & 1225 & 70.78 & 7606 & 28.54 & 12930 & 12.81 & 14410 & 49.76 & 9042.8 & -0.01 & -0.12  \\
    \textit{+ O1-Pruner-4.0} & 91.21 & 1500 & 71.88 & 7848 & 30.63 & 12669 & 12.62 & 14547 & 51.58 & 9141.0 & 0.12 & 0.12  \\
    \textit{+ ShorterBetter} & 84.77 & 200 & 74.84 & 6068 & 27.08 & 11714 & 11.62 & 12908 & 49.58 & 7722.4 & 0.10 & -0.03 \\
    \textit{+ ACPO} & 91.02 & 1282 & 71.25 & 7251 & 30.42 & 12367 & 12.62 & 14189 & 51.33 & 8772.2 & 0.14 & 0.14 \\
    \rowcolor{blue!8} \textit{+ CRT (ours)} & 93.16 & 1582 & 74.38 & 6934 & 28.54 & 12674 & 13.63 & 13956 & 52.43 & 8786.4 & 0.21 & 0.21 \\
\specialrule{.2em}{.1em}{.1em}
\end{tabularx}
\end{table}

\begin{table}[h]
\footnotesize
\centering
\caption{Evaluation results using Qwen3-4B as the base model. Full dataset-wise results are reported in \cref{table:ax-2}.}
\label{table:main-3}
\begin{tabularx}{\columnwidth}{lXXXX}
\specialrule{.2em}{.1em}{.25em}
\textbf{Method} & $\overline{A}$  & $\overline{L}$ & AES$_1$  & AES$_2$ \\
\midrule
    \textbf{Qwen3-4B} & 65.77 & 2472.8 & 0 & 0 \\
    \textit{+ ThinkPrune-500} & 64.96 & 2109.8 & 0.09 & 0.02 \\
    \textit{+ ThinkPrune-1k} & 64.46 & 2006.9 & 0.09 & -0.01 \\
    \textit{+ O1-Pruner-4.0} & 62.66 & 1394.3 & 0.2 & -0.04 \\
    \textit{+ ShorterBetter} & 61.22 & 1343.4 & 0.11 & -0.23 \\
    \textit{+ ACPO}  & 63.12 & 1675.6 & 0.12 & -0.08 \\
    \rowcolor{blue!8} \textit{+ CRT (ours)} & 62.99 & 1361.4 & 0.24 & 0.03 \\
\specialrule{.2em}{.1em}{.1em}
\end{tabularx}
\end{table}

We report the in-domain evaluation results for DeepSeek-1.5B in   \cref{table:main-1} and the out-domain results in \cref{table:main-2}. We report Qwen3-4B results in \cref{table:main-3}. Across all settings, CRT achieves the highest AES scores among the evaluated baselines, indicating a superior ability to balance accuracy and reasoning efficiency. Moreover, CRT preserves the original model's accuracy across most tasks, and in several cases yields measurable improvements. As shown by the training trajectories in \cref{fig:main-training-ckp} (left), CRT consistently reduces average response length throughout training while largely maintaining task performance in both domains. In contrast, ThinkPrune exhibits substantial performance degradation beginning in the mid-training phase. These training-time dynamics directly account for the observed differences in final inference performance.

\subsection{Analyzing Redundancy Beyond Average Length}
While average response length is a useful coarse indicator of reasoning efficiency, it does not fully capture qualitative differences in how different models reason. In particular, two responses of comparable length may differ substantially in informational density, with one exhibiting repetitive or self-referential patterns (e.g., verification loops or self-doubt statements) that do not contribute to problem solving. Motivated by this observation, we move beyond length-based metrics and analyze changes in \emph{language redundancy} across training checkpoints and baselines.

Qualitative inspection at the sample level reveals that earlier checkpoints frequently contain self-checking and verification loops, where later results could be reiterated multiple times throughout the reasoning trace. Whether such patterns are directly correlated with correctness remains unclear; however, they suggest that efficiency gains may arise not only from shorter outputs but also from reduced redundancy within the text. Although one could assess redundancy using semantic judges (e.g., LLM-based evaluators), we instead focus on deterministic, text-based metrics that can be computed directly and reproducibly.

\begin{table}[h]
\footnotesize
\centering
\caption{Evaluation on answer redundancy using GSM8K. }
\label{table:redundancy-1}
\begin{tabularx}{\columnwidth}{lXXX}
\specialrule{.2em}{.1em}{.25em}
\textbf{Method} &  $\bar R_{\text{zip}}$ ($\uparrow$) & $\bar R_{\text{zip}}^{\text{correct}}$ ($\uparrow$) & $\bar R_{\text{zip}}^{\text{wrong}}$ ($\uparrow$) \\
\midrule
    \textbf{DeepSeek-1.5B} & 0.356 & 0.364 & 0.323  \\
    \textit{+ Thinkprune-1k} & 0.388 & 0.395 &  0.363 \\
    \textit{+ O1-Pruner-4.0} & 0.372 & 0.381 & 0.335 \\
    \textit{+ ACPO} & 0.397 & 0.405 & 0.366  \\
    \rowcolor{blue!8} \textit{+ CRT (ours)} & 0.423 & 0.433 & 0.383 \\
\specialrule{.2em}{.1em}{.1em}
\end{tabularx}
\end{table}

\paragraph{Compression Ratio as a Redundancy Metric.}
Redundant text is inherently more compressible. Let $\text{zip}(\cdot)$ denote a deterministic compression algorithm (e.g., gzip), and let $|\cdot|$ denote the byte length of a rollout-level response, including chain-of-thought tokens. We define the compression ratio as
$R_{\text{zip}}(y)=\frac{|\text{zip}(y)|}{|y|}$
where a lower value of $R_{\text{zip}}$ indicates greater internal redundancy. 
We report redundancy statistics across baselines in \cref{table:redundancy-1}, including the average compression ratio $R_{\text{zip}}$ over all responses, as well as its correctness-conditional variants $R_{\text{zip}}^{\text{correct}}$, $R_{\text{zip}}^{\text{wrong}}$. CRT achieves the highest compression ratio, indicating substantially reduced internal redundancy. Notably, although \cref{table:main-1} shows that CRT attains a comparable average output length to ThinkPrune-1k, \cref{table:redundancy-1} reveals a markedly lower redundancy level. This discrepancy demonstrates that average length alone fails to capture structural redundancy in generated reasoning, underscoring the need for redundancy-aware metrics beyond length.

\subsection{Controlling Explanation Length with Correctness Guarantees}

\begin{figure}[h]
  \centering
\includegraphics[width=0.85\linewidth]{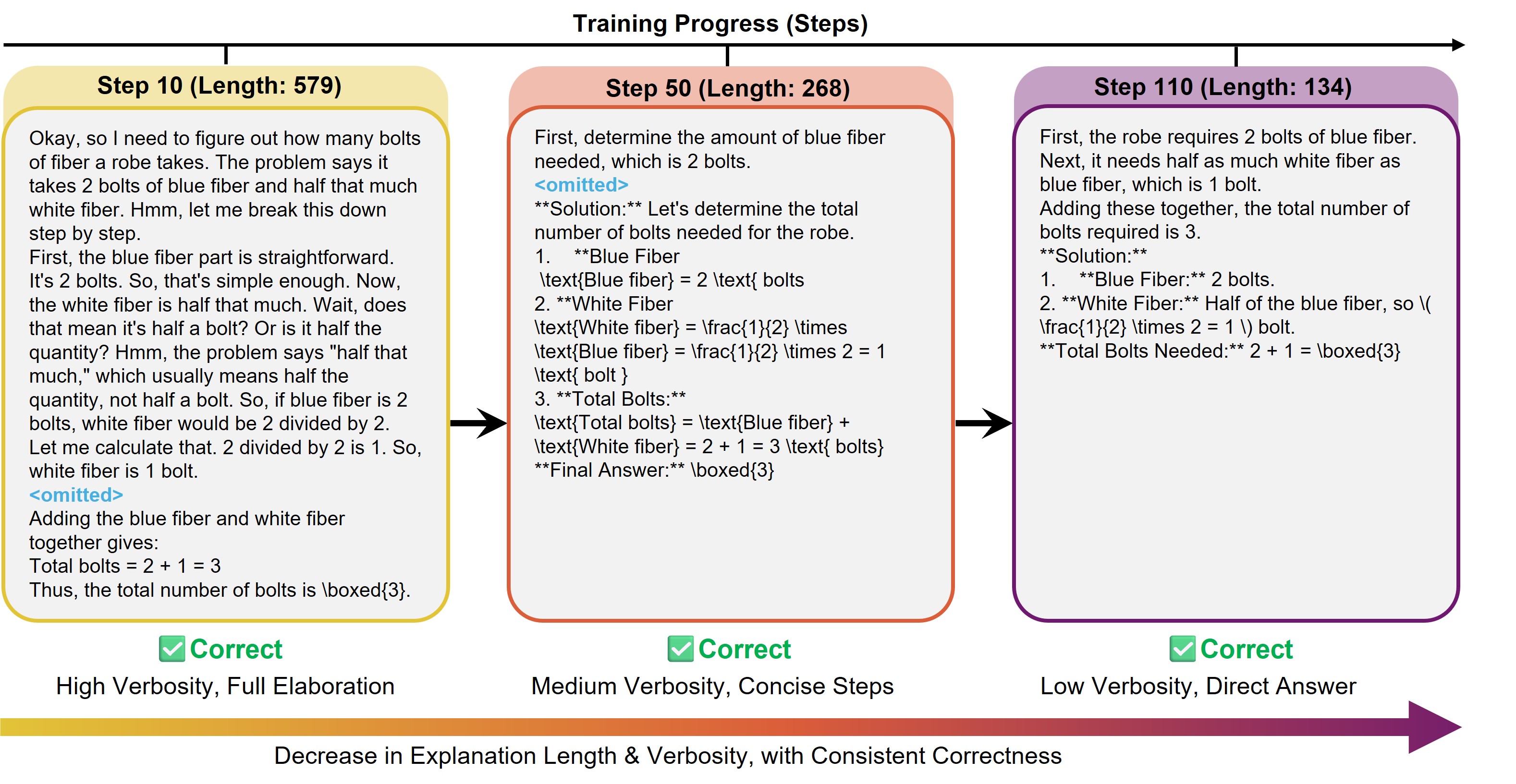}
\caption{Controlling explanation length while preserving correctness across training checkpoints.}
  \label{fig:explanation-1}
\end{figure}

A distinctive advantage of our method is its ability to provide fine-grained, correctness-preserving control over the level of explanation detail. Unlike prior approaches that manipulate the trade-off between correctness and length, our procedure yields a sequence of intermediate checkpoints that span a spectrum of explanation lengths while maintaining original answer quality. This produces a model family that users can query at different levels of verbosity, ranging from concise answers to fully elaborated solutions without sacrificing correctness.
To illustrate this property, we visualize the rollout behavior of intermediate checkpoints on the in-domain GSM8K dataset in \cref{fig:main-training-ckp} (Right).
We further present a qualitative case study in \cref{fig:explanation-1}, which tracks the evolution of a single GSM8K example across intermediate training checkpoints. Despite substantial differences in explanation length and structure, all checkpoints produce the same correct final answer. Earlier checkpoints generate detailed, self-verifying reasoning traces, whereas later checkpoints converge to increasingly compact and direct solutions. This progression highlights a key strength of our approach: explanation length can be continuously adjusted through training progress, rather than discretely tuned via heuristics, while preserving correctness throughout.

\subsection{Ablation Study}
We evaluate the effectiveness of the proposed two-stage training scheme by comparing full CRT against a single-stage variant that follows only Stage~I for the same number of training steps, denoted as \textit{CRT (1s)}. In addition, we include a primal-dual baseline, denoted as \textit{PD}, which jointly optimizes accuracy and length within a single-stage primal-dual formulation. As shown in \cref{table:ablation-1}, \textit{CRT (1s)} consistently underperforms the full two-stage CRT across both in-domain and out-domain evaluations in terms of AES scores. Notably, it exhibits degraded accuracy and longer length on in-domain data, indicating that aggressively minimizing length alone is insufficient to recover reasoning quality once compression saturates. In contrast, the second stage of CRT effectively restores accuracy without substantially increasing response length, leading to improved accuracy-efficiency trade-off. This effect is especially pronounced under AES$_2$, which heavily penalizes accuracy degradation, highlighting the importance of the Stage~II refinement in preserving correctness while maintaining the efficiency profile discovered in Stage~I.

\begin{table}[h]
\footnotesize
\centering
\caption{Evaluation of CRT variants. DeepSeek-1.5B is used as the base model. Detailed per-dataset results are reported in \cref{table:ax-1}.}
\label{table:ablation-1}
\begin{tabularx}{\columnwidth}{lXXXX}
\specialrule{.2em}{.1em}{.25em}
\textbf{Method} & $\overline{A}$  & $\overline{L}$ & AES$_1$  & AES$_2$ \\
\midrule
\textbf{In-domain} \\
    \rowcolor{blue!8} \textit{CRT}& 85.35 & 2499.2 & 0.29 & 0.29 \\
    \textit{CRT (1s)} & 84.52 & 2549.6 & 0.24 & 0.22 \\
    \textit{PD} & 82.28 & 2461 & 0.13 & -0.02 \\
\midrule
\textbf{Out-domain} \\
    \rowcolor{blue!8} \textit{CRT} & 52.43 & 8786.4 & 0.21 & 0.21 \\
    \textit{CRT (1s)} & 52.03 & 8613 & 0.20 & 0.20 \\
    \textit{PD} & 50.1 & 8607.4 & 0.06 & -0.02 \\
\specialrule{.2em}{.1em}{.1em}
\end{tabularx}
\end{table}

\section{Conclusion}
In this work, we revisit the problem of overthinking in LRMs from a principled optimization perspective. While prior approaches often rely on heuristic reward shaping, our analysis highlights that such designs struggle to reliably balance brevity and correctness and are prone to unstable training dynamics. To address this, we introduce \textbf{Constraint-Rectified Training (CRT)}, a reference-guarded post-training framework that minimizes reasoning length while preserving solution quality relative to a frozen reference policy. CRT alternates between pruning redundant reasoning and rectifying accuracy only when performance falls below the reference, yielding a stable and interpretable objective without brittle threshold tuning. CRT further admits a two-stage extension that first discovers the shortest reliable reasoning patterns and then refines accuracy under a learnt length budget, preventing the re-emergence of verbose CoT. Empirically, CRT consistently reduces token usage while maintaining answer quality, improves internal reasoning efficiency beyond length, and naturally produces a spectrum of verbosity-controlled checkpoints without retraining.

\FloatBarrier

\bibliographystyle{abbrvnat}
\bibliography{ref}

\appendix
\section{Hardware Details}
\label{sec:appendix-hardware}
All baselines and our proposed method were trained on systems with the following hardware configuration:
\begin{itemize} 
    \item Operating system: Red Hat Enterprise Linux (RHEL) 9
    \item Type of CPU: Intel Xeon Platinum 8470 (Sapphire Rapids, 52 cores) running at 2.0 GHz
    \item Type of GPU: NVIDIA H100 "Hopper" with 94 GB memory
\end{itemize}

All model inference and evaluation were conducted on separate systems with the following hardware configuration:
\begin{itemize} 
    \item Operating system: Red Hat Enterprise Linux (RHEL) 9
    \item Type of CPU: AMD EPYC 7H12 (64 cores) running at 2.6 GHz 
    \item Type of GPU: NVIDIA A100 "Ampere" with 40 GB memory
\end{itemize}

\section{Additional Details}

\paragraph{Length Normalization.} We implement response-length normalization using a per-prompt standardization scheme: 
\begin{equation}
\ell_{\mathrm{norm}}(y)
=
\sigma\!\left(
\frac{\mathrm{LEN}(y)
-
\mathbb{E}_{y \sim \pi_\theta(\cdot \mid x)}\!\left[\mathrm{LEN}(y)\right]}
{\sqrt{
\mathrm{Var}_{y \sim \pi_\theta(\cdot \mid x)}\!\left[\mathrm{LEN}(y)\right]
}}
\right).
\label{eq:ax-setting-lennorm}
\end{equation}
where the expectation and variance are computed over responses sampled from the current policy. This normalization preserves the relative ordering of response lengths while making length penalties comparable across prompts with different intrinsic difficulty.

\paragraph{Accuracy-Efficiency Score (AES).} 
\label{eq:ax-aes}
This metric was first introduced in O1-Pruner \citep{luo2025o1}, which jointly considers the relative change in accuracy ($\Delta A$) and response length ($\Delta L$), defined as
\[
\Delta A = \frac{A_{\text{model}} - A_{\text{base}}}{A_{\text{base}}}, 
\qquad
\Delta L = \frac{L_{\text{base}} - L_{\text{model}}}{L_{\text{base}}},
\]
where $\Delta L > 0$ indicates a shorter response than the baseline.
The AES is computed as
\[
\mathrm{AES} =
\begin{cases}
\alpha \Delta L + \beta |\Delta A|, & \Delta A \ge 0,\\[4pt]
\alpha \Delta L - \gamma |\Delta A|, & \Delta A < 0,
\end{cases}
\]
where $\alpha$ controls the sensitivity to efficiency gain, while $\beta$ and $\gamma$ govern the reward and penalty magnitudes for accuracy gain or loss, respectively.
We report two variants: $\mathrm{AES}_1$ with $(\alpha,\beta,\gamma) = (1,3,5)$ \citep{luo2025o1} and $\mathrm{AES}_2$ with $(\alpha,\beta,\gamma) = (1,3,10)$ \citep{li2025drpo}.

\section{Additional Results}
We present the full Qwen3-4B results in \cref{table:ax-2}, and report the full CRT ablation study in \cref{table:ax-1}.

\begin{table*}[ht]
\scriptsize
\centering
\caption{Full results of experiments based on Qwen3-4B.}
\label{table:ax-2}
\begin{tabularx}{\textwidth}{lXXXXXXXXXXXXXXXX}\specialrule{.2em}{.1em}{.25em}
\textbf{Method} & \multicolumn{2}{c}{\textbf{GSM8K}} & \multicolumn{2}{c}{\textbf{MATH500}} & \multicolumn{2}{c}{\textbf{SAT MATH}} & \multicolumn{2}{c}{\textbf{AMC23}} & \multicolumn{2}{c}{\textbf{AIME24}} & \multicolumn{2}{c}{\textbf{Olymp.}} & \multicolumn{2}{c}{\textbf{Avg}} & \multicolumn{2}{c}{\textbf{Metrics}} \\
\cmidrule(lr){2-3} \cmidrule(lr){4-5} \cmidrule(lr){6-7} \cmidrule(lr){8-9} \cmidrule(lr){10-11} \cmidrule(lr){12-13} \cmidrule(lr){14-15} \cmidrule(lr){16-17} 
 & Acc & Len & Acc & Len & Acc & Len & Acc & Len & Acc & Len & Acc & Len & $\overline{A}$ & $\overline{L}$ & AES$_1$ & AES$_2$ \\
\midrule
    \textbf{Qwen3-4B}  & 92.2 & 310 & 85.62 & 1113 & 99.02 & 1260 & 69.38 & 1895 & 24.38 & 6382 & 24 & 3876 & 65.77 & 2472.8 & 0 & 0 \\
\midrule
    \textit{+ ThinkPrune-500} & 91.45 & 281 & 84.53 & 1253 & 99.22 & 772 & 67.97 & 1806 & 22.71 & 5332 & 23.88 & 3214 & 64.96 & 2109.8 & 0.09 & 0.02 \\
    \textit{+ ThinkPrune-1k}& 90.32 & 287 & 84.78 & 1167 & 98.83 & 1004 & 66.09 & 1631 & 23.33 & 4879 & 23.38 & 3074 & 64.46 & 2006.9 & 0.09 & $\text{-0.01}$ \\
    \textit{+ O1-Pruner-4.0}  & 90.2 & 186 & 82.12 & 849 & 98.05 & 287 & 63.44 & 1459 & 19.58 & 3222 & 22.56 & 2364 & 62.66 & 1394.3 & 0.2 & $\text{-0.04}$ \\
    \textit{+ ShorterBetter} & 90.16 & 133 & 80.28 & 598 & 96.48 & 1092 & 62.19 & 1064 & 16.04 & 3235 & 22.19 & 1939 & 61.22 & 1343.4 & 0.11 & $\text{-0.23}$ \\
    \textit{+ ACPO}  & 89.98 & 196 & 83.38 & 755 & 99.22 & 549 & 62.81 & 1285 & 21.25 & 4586 & 22.06 & 2683 & 63.12 & 1675.6 & 0.12 & $\text{-0.08}$ \\
    \textit{+ CRT (ours)} & 90.2 & 222 & 82.94 & 790 & 98.83 & 318 & 64.06 & 1387 & 20.21 & 3264 & 21.69 & 2188 & 62.99 & 1361.4 & 0.24 & 0.03 \\
\specialrule{.2em}{.1em}{.1em}
\end{tabularx}
\end{table*}

\begin{table*}[ht]
\scriptsize
\centering
\caption{Full results of CRT ablation study on all testing datasets.}
\label{table:ax-1}
\begin{tabularx}{\textwidth}{lXXXXXXXXXXXXXXXX}
\specialrule{.2em}{.1em}{.25em}
\textbf{Method} & \multicolumn{2}{c}{\textbf{GSM8K}} & \multicolumn{2}{c}{\textbf{MATH500}} & \multicolumn{2}{c}{\textbf{SAT}} & \multicolumn{2}{c}{\textbf{AMC23}} & \multicolumn{2}{c}{\textbf{AIME24}} & \multicolumn{2}{c}{\textbf{Olymp.}} & \multicolumn{2}{c}{\textbf{Avg}} & \multicolumn{2}{c}{\textbf{Metrics}} \\
\cmidrule(lr){2-3} \cmidrule(lr){4-5} \cmidrule(lr){6-7} \cmidrule(lr){8-9} \cmidrule(lr){10-11} \cmidrule(lr){12-13} \cmidrule(lr){14-15} \cmidrule(lr){16-17} 
 & Acc & Len & Acc & Len & Acc & Len & Acc & Len & Acc & Len & Acc & Len & $\overline{A}$ & $\overline{L}$ & AES$_1$ & AES$_2$ \\
\midrule
    \textit{DeepSeek-1.5B} & 84.43 & 1869 & 85.19 & 4987 & 89.84 & 1373 & 69.84 & 8634 & 30.42 & 14044 & 13.25 & 15527 & 62.16 & 7738.9 & 0 & 0 \\
\midrule
    \textit{CRT} & 83.23 & 1102 & 85.81 & 3998 & 93.16 & 1214 & 72.66 & 6982 & 29.17 & 12426 & 13.12 & 13831 & 62.86 & 6591.8 & 0.18 & 0.18 \\
    \textit{CRT (1s)} & 84.64 & 1172 & 86.06 & 3826 & 93.16 & 1582 & 74.38 & 6934 & 28.54 & 12674 & 13.63 & 13956 & 63.4 & 6690.7 & 0.20 & 0.20 \\
    \textit{PD} & 79.24 & 975 & 85.31 & 3947 & 88.67 & 803 & 71.72 & 7337 & 27.5 & 12540 & 12.5 & 13750 & 60.82 & 6558.6 & 0.04 & $\text{-0.06}$ \\
\specialrule{.2em}{.1em}{.1em}
\end{tabularx}
\end{table*}

\begin{algorithm}[H]
\caption{Primal-Dual Finetuning (PDO~\cite{chow2018risk} variant)}
\small
\label{algo:main-pd}
\begin{algorithmic}[1]
\STATE \textbf{Input:} Initial LM parameters $\theta_0$, reference policy $\pi_{\mathrm{ref}}$,
dual variable $\lambda_0 \ge 0$, learning rates $\eta_\theta, \eta_\lambda$,
tolerance $\varepsilon$, total steps $T$.
\FOR{step $t = 0, 1, \dots, T-1$}
    \STATE Sample minibatch $\{x_i\}_{i=1}^B \sim \rho$.
    \STATE Generate $y_i \sim \pi(y \mid x_i, \theta_t)$ and $y^{\mathrm{ref}}_i \sim \pi_{\mathrm{ref}}(y \mid x_i)$.

    \STATE Estimate minibatch accuracies:
    
    \(
    \begin{aligned}
    \hat{A}_t
    &= \frac{1}{B}\sum_{i=1}^B \mathds{1}\{y_i = y^*(x_i)\}, \\
    \hat{A}^{\mathrm{ref}}
    &= \frac{1}{B}\sum_{i=1}^B \mathds{1}\{y^{\mathrm{ref}}_i = y^*(x_i)\}.
    \end{aligned}
    \)

    \STATE Compute constraint residual:
    \[
    \hat{g}_t \leftarrow (\hat{A}^{\mathrm{ref}} - \varepsilon) - \hat{A}_t .
    \]

    \STATE Compute Lagrangian estimator:
    \[
    \begin{aligned}
    \hat{\mathcal{L}}(\theta_t, \lambda_t)
    &= \frac{1}{B}\sum_{i=1}^B \ell_{\mathrm{norm}}(y_i)
    \;+\; \lambda_t \, \hat{g}_t .
    \end{aligned}
    \]

    \STATE /* Primal update on policy: */
    \[
    \theta_{t+1} \leftarrow \theta_t - \eta_\theta \nabla_\theta \hat{\mathcal{L}}(\theta_t, \lambda_t).
    \]

    \STATE /* Dual update on constraint: */
    \[
    \lambda_{t+1} \leftarrow \big[\lambda_t + \eta_\lambda \, \hat{g}_t \big]_+.
    \]
\ENDFOR
\end{algorithmic}
\end{algorithm}

 We provide the full text for \cref{fig:explanation-1}:
 
\begin{promptbox}[Step 10 (Length: 579)]{yellow}
\begin{ttfamily}
    Okay, so I need to figure out how many bolts of fiber a robe takes. The problem says it takes 2 bolts of blue fiber and half that much white fiber. Hmm, let me break this down step by step. 
    
    First, the blue fiber part is straightforward. It's 2 bolts. So, that's simple enough. Now, the white fiber is half that much. Wait, does that mean it's half a bolt? Or is it half the quantity? Hmm, the problem says "half that much," which usually means half the quantity, not half a bolt. So, if blue fiber is 2 bolts, white fiber would be 2 divided by 2. Let me calculate that. 2 divided by 2 is 1. So, white fiber is 1 bolt. 
    
    Now, to find the total number of bolts, I need to add the blue fiber and white fiber together. So, that's 2 bolts plus 1 bolt. Let me do the addition: 2 + 1 equals 3. So, altogether, the robe takes 3 bolts of fiber. 
    
    Wait, just to double-check, sometimes in math problems, especially with word problems, the wording can be tricky. The problem says "half that much." So, is it 50\% more than half of the blue fiber? No, that doesn't make sense because it's saying half that much, not half more. So, half of 2 is 1. That still seems correct. 
    
    Alternatively, if someone misreads it as half as much, that would be 1 bolt, but the problem says half that much, which is 1. So, yes, 1 bolt white fiber. Adding them together gives 3 total bolts. 
    
    I don't think there's any other way to interpret "half that much." If it were half of the blue fiber, that would be 1, but half that much would mean half the amount, so half of 2 is 1. Either way, it's 1. So, adding them gives 3. 
    
    I think that's solid. So, the total number of bolts is 3. 
    
    **Final Answer**
    The robe takes a total of \boxed{3} bolts of fiber.

    The robe takes 2 bolts of blue fiber. The white fiber is half that much, which is calculated as follows:
    
    \[
    \text{White fiber} = \frac{2}{2} = 1 \text{ bolt}
    \]
    
    Adding the blue fiber and white fiber together gives:
    
    \[
    \text{Total bolts} = 2 + 1 = 3
    \]
    
    Thus, the total number of bolts is \boxed{3}.
\end{ttfamily}
\end{promptbox}

\begin{promptbox}[Step 50 (Length: 268)]{orange}
\begin{ttfamily}
    First, determine the amount of blue fiber needed, which is 2 bolts.
    
    Next, calculate the white fiber required. It's half of the blue fiber, so 2 divided by 2 equals 1 bolt.
    
    Finally, add the two amounts together to find the total number of bolts: 2 plus 1 equals 3 bolts.

    **Solution:**
    
    Let's determine the total number of bolts needed for the robe.
    
    1. **Blue Fiber:**
       \[
       \text{Blue fiber} = 2 \text{ bolts}
       \]
    
    2. **White Fiber:**
       \[
       \text{White fiber} = \frac{1}{2} \times \text{Blue fiber} = \frac{1}{2} \times 2 = 1 \text{ bolt}
       \]
    
    3. **Total Bolts:**
       \[
       \text{Total bolts} = \text{Blue fiber} + \text{White fiber} = 2 + 1 = 3 \text{ bolts}
       \]
    
    **Final Answer:**
    \[
    \boxed{3}
    \]
\end{ttfamily}
\end{promptbox}

\begin{promptbox}[Step 110 (Length: 134)]{purple}
\begin{ttfamily}
    First, the robe requires 2 bolts of blue fiber.
    
    Next, it needs half as much white fiber as blue fiber, which is 1 bolt.
    
    Adding these together, the total number of bolts required is 3.

    **Solution:**
    
    1. **Blue Fiber:** 2 bolts.
    
    2. **White Fiber:** Half of the blue fiber, so \( \frac{1}{2} \times 2 = 1 \) bolt.
    
    **Total Bolts Needed:**  
    \( 2 + 1 = \boxed{3} \)
\end{ttfamily}
\end{promptbox}

\section{Case Study of Inference Behavior}
\begin{figure}[p]
  \centering
\includegraphics[width=0.51\linewidth]{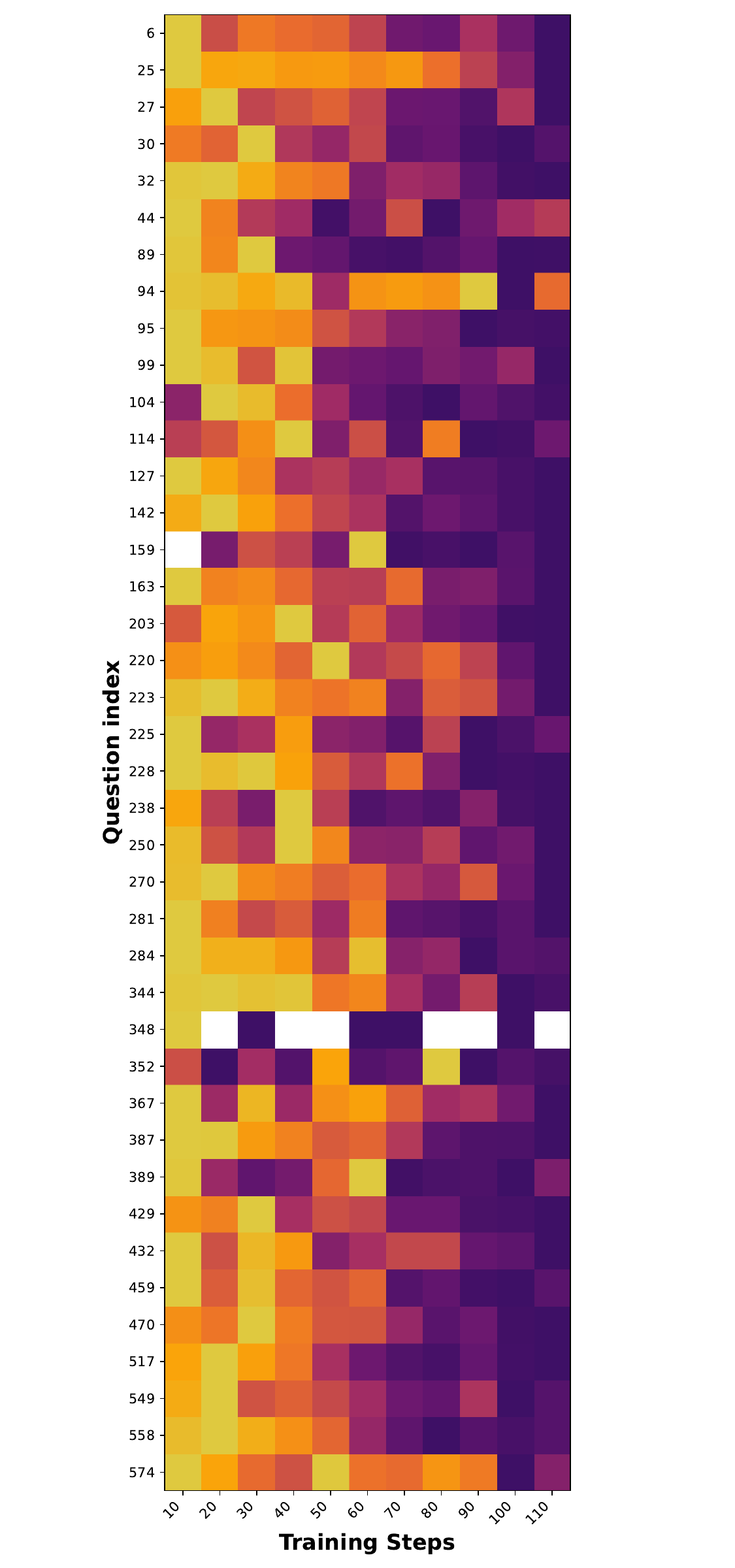}
\hspace{-0.1\linewidth}
\includegraphics[width=0.51\linewidth]{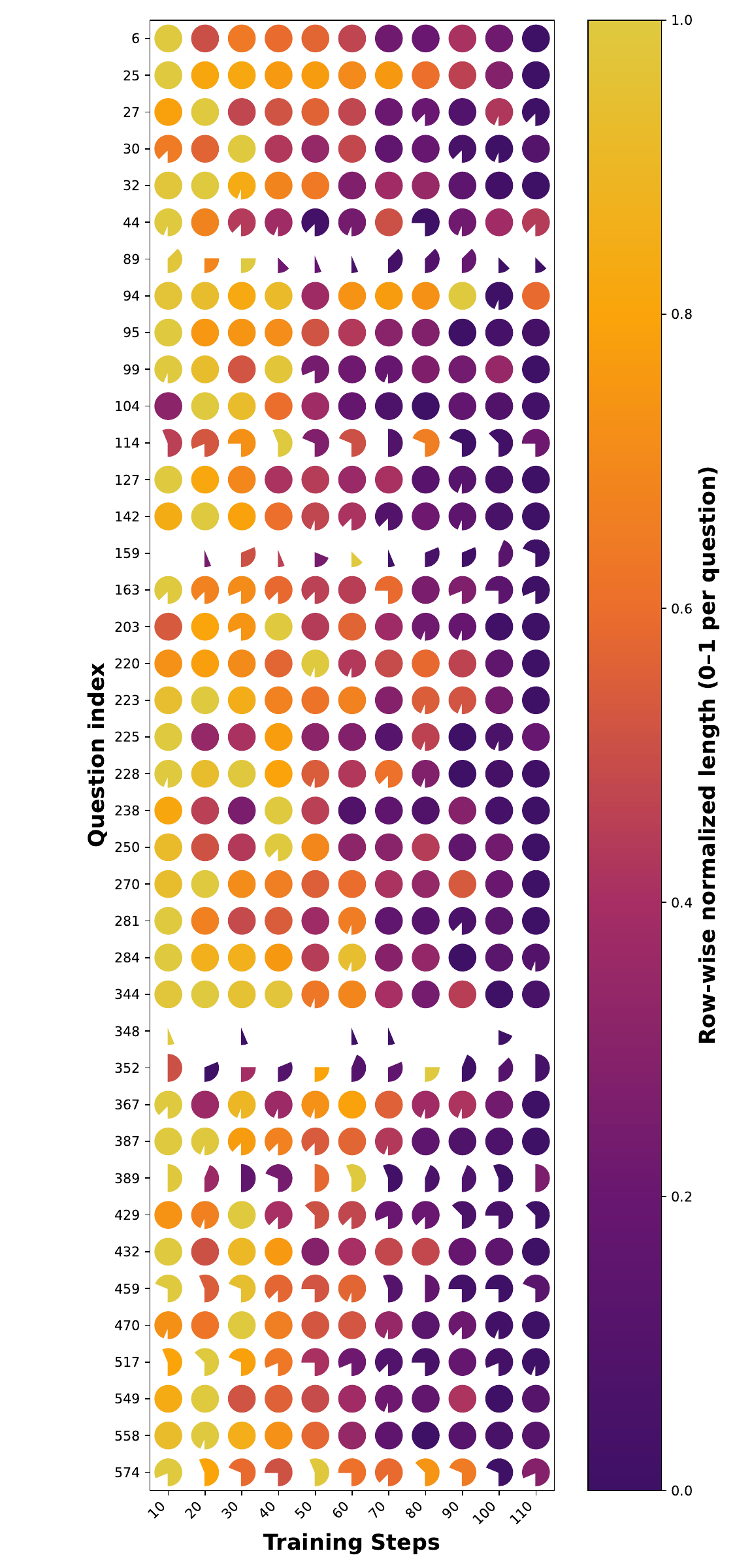}
\caption{Case study of GSM8K inference behavior. Each row corresponds to a question, and each column to an intermediate model checkpoint generated during CRT training. Color intensity encodes the response length (brighter = longer; darker = shorter). Circle fill represents accuracy: a fully filled circle marks a correct prediction, whereas a partially filled circle reflects partial correctness.}
  \label{fig:appendix-casestudy-1}
\end{figure}

We present the full case study corresponding to the right panel of  \cref{fig:main-training-ckp} in \cref{fig:appendix-casestudy-1}. The results show that response length is consistently reduced over training for most samples, while the majority of questions preserve performance or even exhibit accuracy improvements.

\end{CJK*}
\end{document}